\documentclass[10pt,twocolumn,letterpaper]{article}

\usepackage[pagenumbers]{cvpr} % To force page numbers, e.g. for an arXiv version
\usepackage{graphicx}%
\usepackage{multirow}%
\usepackage{amsmath,amssymb,amsfonts}%
\usepackage{amsthm}%
\usepackage{mathrsfs}%
\usepackage[title]{appendix}%
\usepackage{textcomp}%
\usepackage{manyfoot}%
\usepackage{booktabs}%
\usepackage{algorithm}%
\usepackage{algorithmicx}%
\usepackage{algpseudocode}%
\usepackage{listings}%
\usepackage{url}            % simple URL typesetting
\usepackage{nicefrac}       % compact symbols for 1/2, etc.
\usepackage[symbol]{footmisc}
\usepackage{microtype}      % microtypography
\usepackage{multicol}
\usepackage{stfloats}
\usepackage{caption}
\usepackage{epsfig}
\usepackage{lipsum}  % for generating dummy text
\usepackage[accsupp]{axessibility}
\usepackage[dvipsnames]{xcolor}

\definecolor{cvprblue}{rgb}{0.21,0.49,0.74}
\usepackage[pagebackref,breaklinks,colorlinks,citecolor=cvprblue]{hyperref}

\def\confName{CVPR}
\def\confYear{2024}

\title{REACTO: Reconstructing Articulated Objects from a Single Video}
\author{
Chaoyue Song\textsuperscript{1,2}, Jiacheng Wei\textsuperscript{1, \dag}, Chuan Sheng Foo\textsuperscript{2,3}, Guosheng Lin\textsuperscript{1, \dag}, Fayao Liu\textsuperscript{2, \dag}\\
\textsuperscript{1}{Nanyang Technological University},
\textsuperscript{2}{Institute for Inforcomm Research, A*STAR} \\
\textsuperscript{3}{Centre for Frontier AI Research, A*STAR} \\  
\small\textsuperscript{\dag}{Corresponding Authors} \\  
{\tt\small \{chaoyue002@e., jiacheng.wei@, gslin@\}ntu.edu.sg, \{foo\_chuan\_sheng, liu\_fayao\}@i2r.a-star.edu.sg}
}

\begin{document}

% \twocolumn[{%
% \renewcommand\twocolumn[1][]{#1}%
\maketitle

\begin{abstract}
In this paper, we address the challenge of reconstructing general articulated 3D objects from a single video.
Existing works employing dynamic neural radiance fields have advanced the modeling of articulated objects like humans and animals from videos, but face challenges with piece-wise rigid general articulated objects due to limitations in their deformation models. To tackle this, we propose Quasi-Rigid Blend Skinning, a novel deformation model that enhances the rigidity of each part while maintaining flexible deformation of the joints. Our primary insight combines three distinct approaches: 1) an enhanced bone rigging system for improved component modeling, 2) the use of quasi-sparse skinning weights to boost part rigidity and reconstruction fidelity, and 3) the application of geodesic point assignment for precise motion and seamless deformation. Our method outperforms previous works in producing higher-fidelity 3D reconstructions of general articulated objects, as demonstrated on both real and synthetic datasets. Project page: \url{https://chaoyuesong.github.io/REACTO}.
\end{abstract}

\section{Introduction}
\begin{figure}[t]
  \centering
  \includegraphics[scale=0.29]{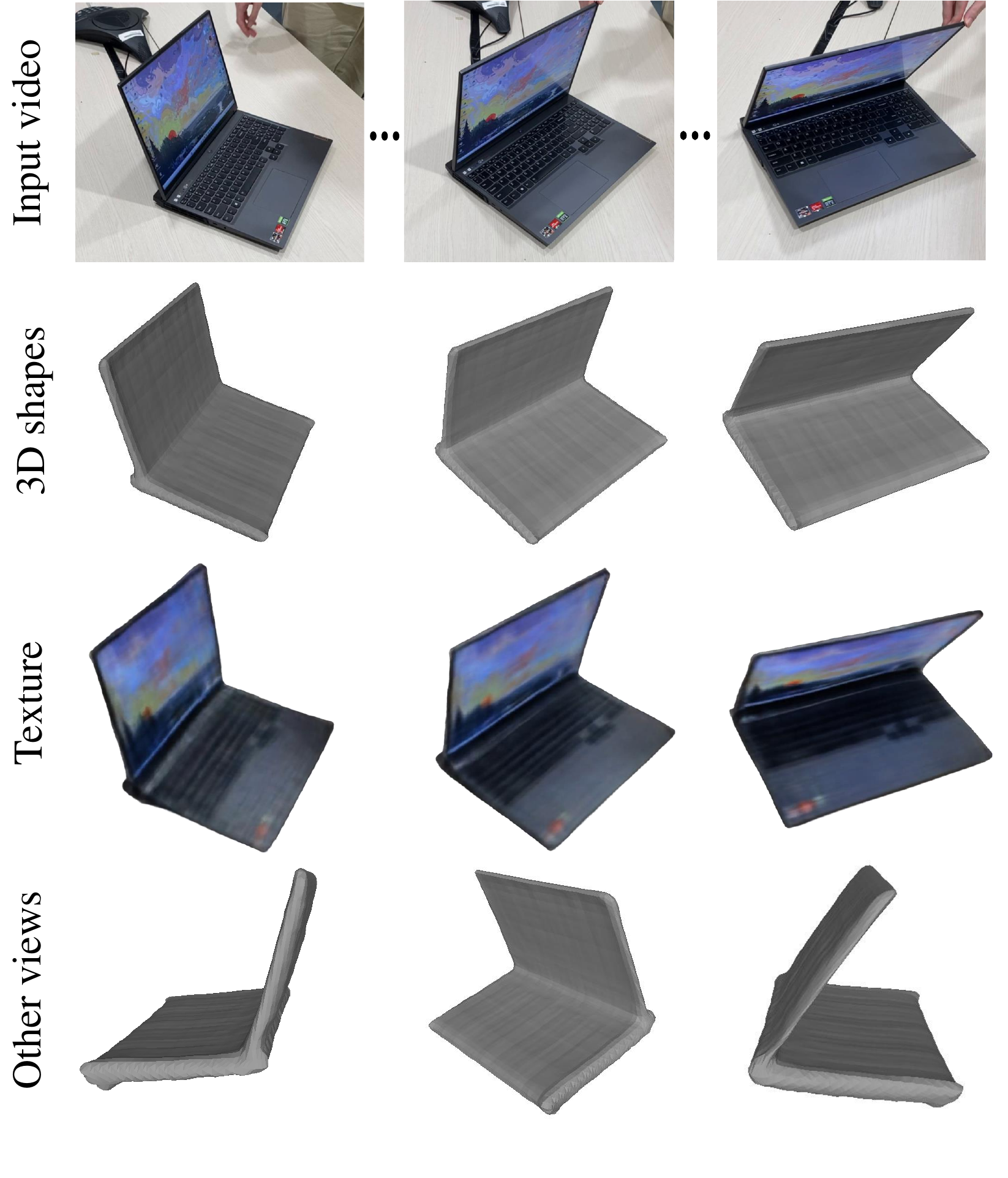}
  \caption{Given a single casual video capturing a piece-wise rigid general articulated object, REACTO can model the 3D shape, texture, and motion. The second row presents shape reconstruction results from reference views, the third row showcases the reconstructed texture, and the fourth row displays the shapes from another view.}
\end{figure}

We focus on reconstructing general articulated objects from a casually captured monocular video, a challenging task that involves creating 3D models from everyday footage and dealing with the complexity of objects with movable parts. Understanding and recognizing the structure of general articulated objects from videos plays a crucial role in various fields, such as robotics, animation, 3D generation \cite{chen2024sculpt3d, wei2023taps3d, chen2023gaussianeditor} virtual reality, and augmented reality. 

Recently, NASAM \cite{wei2022self} introduced a method to learn categories of articulated objects from multi-view images across various articulations. However, this approach necessitates training on several objects within the same category. Another method, PARIS \cite{liu2023paris}, was proposed to learn articulation in a self-supervised manner but relies on multi-view images to provide complete views of the object at different articulations. Consequently, both of these methods face limitations when applied to casually captured everyday videos.

Previous research \cite{weng2022humannerf, guo2023vid2avatar, jiang2022neuman, jiang2022selfrecon} on reconstructing articulated objects from monocular videos has primarily focused on humans and quadrupeds, utilizing readily available parametric models like SMPL \cite{loper2015smpl} and SMAL \cite{zuffi20173d}, while neglecting the diverse range of everyday objects we commonly encounter and use. Non-parametric methods, like BANMo \cite{yang2022banmo}, MoDA \cite{song2023moda}, and PPR\cite{yang2023ppr}, utilizing dynamic volumetric neural radiance fields to model deformable objects. These methods are predominantly optimized for non-rigid, deformable subjects such as humans and animals, whose movable parts such as arms and legs are distinctly separated. However, the rigid movable components of general objects often sit adjacent to each other in their usual poses. For instance, consider the blades of a pair of scissors, which come into close proximity during use. This presents a considerable challenge to the previously mentioned methods with blend skinning techniques used for motion modeling, often leading to incorrect motions and artifacts.

Specifically, BANMo \cite{yang2022banmo} utilizes neural Linear Blend Skinning (LBS) as the deformation model, while LBS is efficient and straightforward, it can sometimes lead to unrealistic deformations, resulting in substantial defects like candy wrapper artifacts and volume loss. MoDA \cite{song2023moda} proposes to use Neural Dual Quaternion Blend Skinning (NeuDBS) to relieve these issues. Although NeuDBS offers improvements in handling rotations and preserving volume, it can still lead to the generation of unsmooth, less refined surfaces. PPR \cite{yang2023ppr} incorporates Dual Quaternion Blend Skinning (DBS) along with novel skin losses and a more stable eikonal loss \cite{gropp2020implicit} to enhance the overall surface smoothness. However, it is observed that surfaces of one moving part tend to tear and get drawn towards another, with visible seam artifacts. Furthermore, the joints appear over-smoothed, leading to a loss of geometric precision. These defects likely arise from inaccurately assigned skinning weights. 

\begin{figure}[t]
  \centering
  \setlength{\abovecaptionskip}{0.cm}
  \includegraphics[scale=0.19]{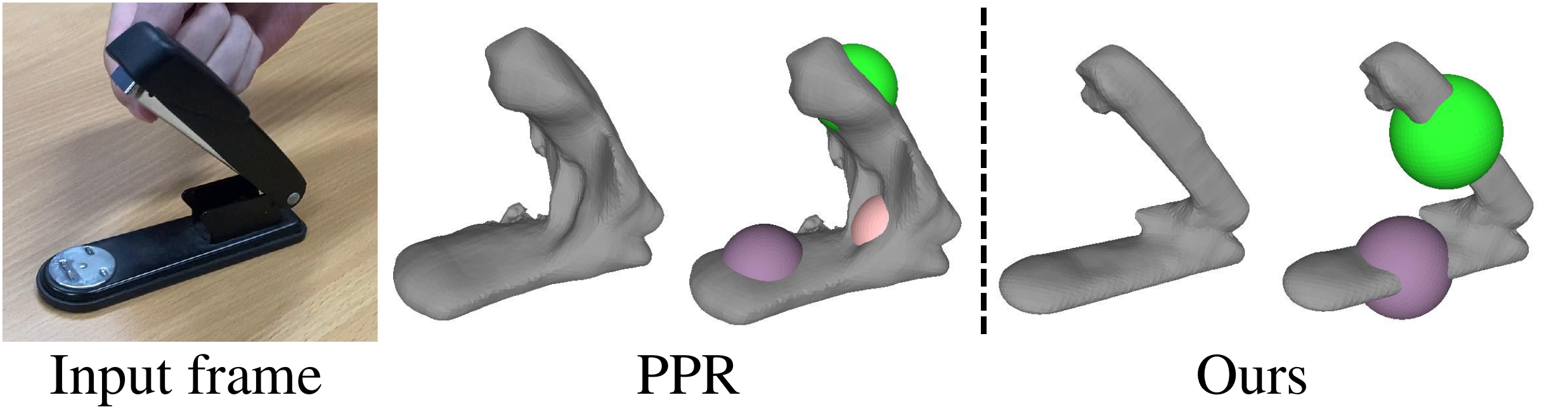}
  \caption{\textbf{Rig on joints vs. rig on bones.} A straightforward approach to control the motion of general articulated objects is to adopt methods \cite{yang2023ppr} used for modeling humans or animals, which typically define the rig based on joints. This design can lead to bending shapes and corrupted motion. In contrast, we propose a novel approach by defining the rig based on bones, enhancing the rigidity and motion integrity of each component.
  } 
  \label{fig2}
\end{figure}

In this work, we present \textbf{REACTO} to REconstruct general ArtiCulaTed Objects from a single casually captured monocular video. Methods with conventional blend skinning techniques, like SMPL \cite{loper2015smpl}, define their rig on the joints, note that some methods like BANMo \cite{yang2022banmo} refer to joints as bones. In this case, it has been observed that the reconstructed shape of each rigid component can be bent by two joints, sometimes leading to seam artifacts. To address this, we define the rig on the bones. As depicted in \Cref{fig2}, our method optimizes the placement of bones to be near the centroid of each component, effectively enhancing the rigidity and motion integrity of these components. 

As discussed previously, the defects also stem from inaccurately assigned skinning weights. For each rigid component, these problems can be addressed by implementing Rigid Skinning (RS), where each vertex is exclusively linked to a single bone. However, RS fails in modeling deformations near joints and can also lead to unwanted discontinuities. To overcome this, we propose Quasi-Rigid Blend Skinning, which merges the rigidity of RS with the flexibility of DBS. Specifically, we optimize the skinning weights on rigid components to be quasi-sparse, minimizing the influence from other bones and ensuring a strong association with their corresponding bone, thus displaying characteristics of rigid skinning. Concurrently, points near joints retain the adaptability inherent in DBS. The accuracy of the commonly used Mahalanobis distance \cite{yang2021lasr, yang2022banmo} is often compromised because its calculation relies on the precision of bone properties, including center, orientation, and scale, all of which are optimized during training. Consequently, we utilize geodesic distance as a more effective measure to jointly ascertain the appropriate corresponding bone for each point or to determine if the point is part of a joint. We demonstrate through experiments that REACTO consistently produces 3D shapes with higher-fidelity details compared to previous state-of-the-art approaches \cite{yang2022banmo, song2023moda, yang2023ppr}.

We summarize our contributions as:
\begin{itemize}
    \item We present REACTO, a novel approach for modeling general articulated 3D objects from single casual videos, without complete views of the objects and any 3D supervision. REACTO demonstrates superior performance over current methods on both real and synthetic datasets.
    \item We redefine the rigging structure in our approach by placing rigs on the bones instead of joints, enhancing the rigidity and motion integrity of each component in general articulated objects.
    \item We propose Quasi-Rigid Blend Skinning (QRBS), a hybrid technique that harmonizes the rigidity of Rigid Skinning with the flexibility of Dual Quaternion Blend Skinning, empowered by quasi-sparse skinning weights, and geodesic point assignment for precise motion reconstruction of general articulated objects.

\end{itemize}

\section{Related Work}
\vspace{-5pt}
\paragraph{Modeling articulated objects.} In the field of computer vision, previous research on articulated deformations has predominantly concentrated on human and animal subjects \cite{loper2015smpl, zuffi20173d, xu2020ghum, romero2022embodied, noguchi2021neural, mihajlovic2021leap, joo2018total, deng2020nasa, song20213d, song2023unsupervised, yang2023attrihuman}, with less attention given to the modeling of general articulated objects that exhibit piece-wise rigidity. Building on the advancements in implicit representations like DeepSDF \cite{park2019deepsdf},  A-SDF \cite{mu2021sdf} models category-level articulation by introducing distinct shape and articulation codes. It integrates joint angles into the shape code, thereby learning to map these angles to their corresponding deformed shapes. ANCSH \cite{li2020category} introduces normalized articulated object coordinate space to model the canonical representations of articulated objects at the category level. CAPTRA \cite{weng2021captra} presents a unified framework for online pose tracking of both rigid and articulated objects from point cloud sequences. Ditto \cite{jiang2022ditto} predicts motion and geometry across object categories using two 3D point clouds, but it struggles with generalizing to new categories and detailed appearance reconstruction. StrobeNet \cite{zhang2021strobenet} extends the previous works to reconstruct articulated objects from multi-view images. However, these methods require ground-truth 3D data for processing or training. CARTO \cite{heppert2023carto} and NASAM \cite{wei2022self} can model articulated objects without 3D ground truth but require category-specific training data with multiple objects. PARIS \cite{liu2023paris}, designed for self-supervised learning of articulation, can generalize to new objects but relies on complete multi-view images, limiting its applicability to casually captured videos.
 \vspace{-12pt}
\paragraph{Shape reconstruction from images or videos.} Various methods have been developed to learn 3D reconstruction from images or videos, guided by annotations like 3D key points \cite{kanazawa2018learning, li2020online}, optical flow \cite{wu2021dove}, and semantic mask \cite{goel2020shape, li2020self, ye2021shelf}. However, the models suffer in generalization since they heavily rely on prior shape templates. Neural implicit surface representations \cite{oechsle2021unisurf, saito2019pifu, saito2020pifuhd, wang2021neus, yariv2020multiview} have found extensive use in reconstructing images or videos. Works like \cite{novotny2017learning, henzler2021unsupervised} have focused on reconstructing rigid objects from videos, however, they fall short in modeling articulated and deformable objects. Recent advancements, including LASR \cite{yang2021lasr} and ViSER \cite{yang2021viser}, have made strides in optimizing a single 3D deformable model from a monocular video guided by mask and optical flow, yet the reconstructed motion often presents unrealistic artifacts. Several studies \cite{liu2021neural, peng2021neural, noguchi2021neural, neural-human-radiance-field, yu2021pixelnerf, chen2021mvsnerf, song2022nerfplayer, listreaming, li2022tava, su2021nerf} have explored reconstructing shape and appearance from images or videos relied on neural radiance fields (NeRF) \cite{mildenhall2021nerf}. In this study, we model general articulated objects from a single video, employing a canonical Neural Radiance Field (NeRF) for shape and appearance, coupled with a deformation model that facilitates the transformation of 3D points between observation and canonical spaces.

 \vspace{-12pt}
\paragraph{Neural representations for dynamic scenes.} Several recent studies have focused on developing deformation models that characterize dynamic scenes by transforming 3D points between the observation space and the canonical space. NR-NeRF \cite{tretschk2021non} depicts deformations on non-rigid objects by learning a rigidity network. D-NeRF \cite{pumarola2021d} is designed to transform points to the canonical space by learning a displacement, while NSFF \cite{li2021neural} displaces 3D points utilizing scaled scene flow. Additionally, Nerfies \cite{park2021nerfies} and HyperNeRF \cite{park2021hypernerf} define deformation by employing a learned dense $\mathrm{SE(3)}$ field. These approaches, however, tend to struggle with large motions between foreground objects and their backgrounds. To address these challenges, several works \cite{neural-human-radiance-field, noguchi2021neural, peng2021animatable, weng2022humannerf} employ the parametric 3D human models, such as SMPL \cite{loper2015smpl}, while other methods \cite{liu2021neural, peng2021animatable, peng2021neural} utilize synchronized multi-view video inputs. BANMo \cite{yang2022banmo}, MoDA \cite{song2023moda}, RAC \cite{yang2023reconstructing}, Total-Recon \cite{song2023total} and PPR\cite{yang2023ppr} can reconstruct 3D shapes from casual videos without relying on human or animal models, by adopting linear blend skinning or dual quaternion blend skinning to learn the deformation model. However, these methods often result in notable artifacts when applied to general articulated objects. To solve this problem, we propose quasi-rigid blend skinning (QRBS) to model the motion of general articulated objects.

\begin{figure*}[h]
  \centering
  \setlength{\abovecaptionskip}{0.cm}
  \includegraphics[scale=0.38]{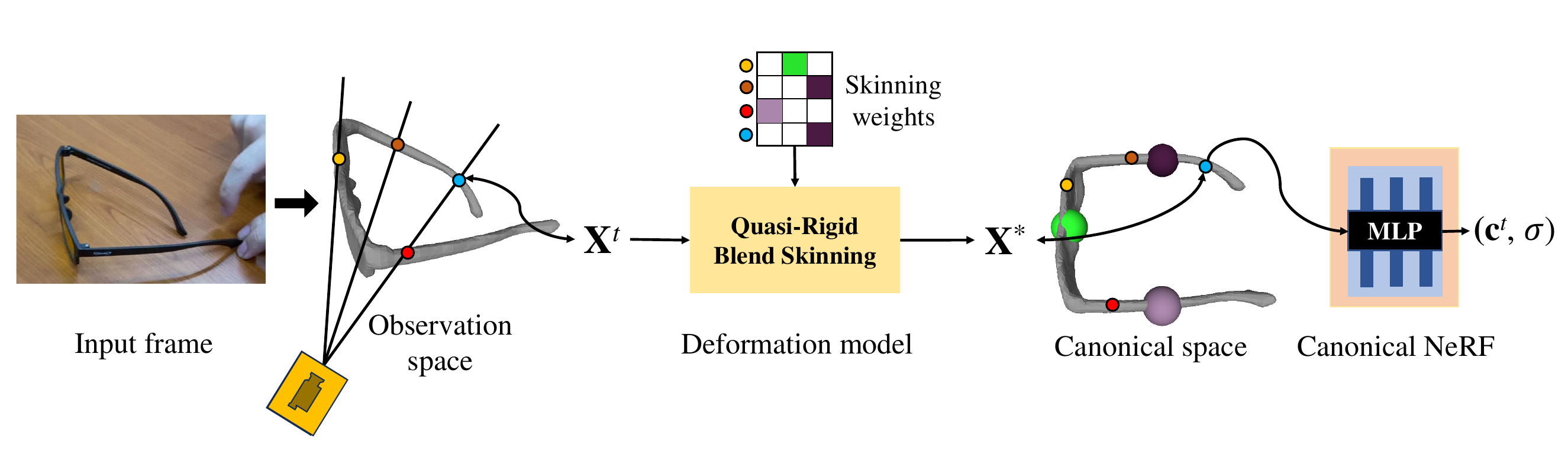}
  \caption{\textbf{The overview of REACTO.} We model an articulated 3D object from a single video using a shape and appearance model based on a canonical Neural Radiance Field (NeRF) and a deformation model for transforming 3D points between the observation space and the canonical space. Instead of linear blend skinning or dual quaternion blend skinning designed for human or animal motion modeling, we propose Quasi-Rigid Blend Skinning (QRBS) as our deformation model, with the learned quasi-sparse skinning weights, to accurately transform $\mathbf{X}^{t}$ from the observation space to $\mathbf{X}^{*}$ in the canonical space. We visualize the 3 bones for glasses in the canonical space. The colors in skinning weights signify the assigned bone for each point.
  } 
  \label{fig_method}
   \vspace{-15pt}
\end{figure*}

\vspace{-5pt}
\section{Method}

The overview of our approach is illustrated in \Cref{fig_method}. In this work, we undertake the task of modeling a 3D articulated object from a single video, employing a canonical Neural Radiance Field (NeRF) as the basis for our shape and appearance model (\Cref{nerf}). Additionally, our approach includes a deformation model (\Cref{deformation}) that transforms 3D points between observation and canonical spaces. Traditional methods like linear blend skinning or dual quaternion blend skinning, typically used for human or animal motion modeling, are inadequate for capturing motion in general articulated objects with multiple rigid components. To overcome this, we introduce Quasi-Rigid Blend Skinning (QRBS) as our deformation model, providing a more apt solution for modeling the motion of such objects. The models are then optimized using volume rendering (\Cref{loss}).

\subsection{Canonical NeRF for shape and appearance}
\label{nerf}
We first define the canonical NeRF \cite{mildenhall2021nerf} to model the shape and appearance of an articulated object. As in BANMo \cite{yang2022banmo}, we learn the color and density of a 3D point $\mathbf{X}^{*} \in \mathbb{R}^{3}$ in the canonical space,
\vspace{-5pt}
\begin{equation}
    \mathbf{c}^{t} = \mathbf{MLP}_{\mathrm{color}}(\mathbf{X}^{*}, \mathbf{D}^{t}, \boldsymbol{\psi}_{a}^{t}),
    \label{color}
\end{equation}
\vspace{-10pt}
\begin{equation}
    \sigma = \Phi_{\beta}(\mathbf{MLP}_{\mathrm{SDF}}(\mathbf{X}^{*})),
    \label{density}
\end{equation}
where $\mathbf{MLP}_{\mathrm{color}}$ and $\mathbf{MLP}_{\mathrm{SDF}}$ are multi-layer perceptron (MLP) networks, $\mathbf{D}^{t} = (\phi^{t}, \theta^{t})$ is the time-varying view direction and $\boldsymbol{\psi}_{a}^{t}$ is a 64-dimensional latent appearance code, serving to encode variations in appearance \cite{martin2021nerf}. To perform volume rendering as \cite{mildenhall2021nerf}, we follow \cite{wang2021neus, yariv2021volume} to use the Cumulative Distribution Function $\Phi_{\beta}(\cdot)$ of the Laplace distribution with zero mean and $\beta$ scale to convert signed distances into density. Here, $\beta$ is a learnable parameter that controls the solidness of the object.

\subsection{Quasi-rigid blend skinning for deformation}
\label{deformation}
With the 3D point $\mathbf{X}^{*}$ in the canonical space and $\mathbf{X}^{t}$ in the observation space, we achieve 3D deformation between them via the deformation model. The canonical-to-observation and observation-to-canonical deformation at time $t$ are denoted as $\mathcal{D}^{t, c\xrightarrow{}o}$ and $\mathcal{D}^{t, o\xrightarrow{}c}$ respectively.

\vspace{-10pt}
\paragraph{Motion representation.} 
For the motion of articulated objects, it encompasses global-level transformations $\mathbf{T}_{\mathrm{global}} \in  \mathrm{SE}(3)$ and object-level articulation $\mathbf{T}_{\mathrm{obj}} \in \mathbb{R}^{8}$ represented by a unit dual quaternion. Given the 3D point $\mathbf{X}^{*}$ in the canonical space and $\mathbf{X}^{t}$ in the observation space, we can deform one to the other via

\begin{equation}
    \mathbf{X}^{t} = \mathcal{D}^{t, c\xrightarrow{}o}(\mathbf{X}^{*}) = \mathbf{T}^{t}_{\mathrm{global}}\mathbf{T}^{t, c\xrightarrow{}o}_{\mathrm{obj}}\mathbf{X}^{*},
\end{equation}
\begin{equation}
    \mathbf{X}^{*} = \mathcal{D}^{t, o\xrightarrow{}c}(\mathbf{X}^{t}) = \mathbf{T}_{\mathrm{obj}}^{t, o\xrightarrow{}c}(\mathbf{T}_{\mathrm{global}}^{t})^{-1}\mathbf{X}^{t},
\end{equation}
where $\mathbf{T}_{\mathrm{global}}$ comprises camera pose transformations $\mathbf{T}_{\mathrm{cam}}$ and root body transformations $\mathbf{T}_{\mathrm{root}}$, both modeled as per-frame SE(3) transformations represented by MLP networks. A detailed introduction to the object-level articulation of general articulated objects will be provided in the following.

\vspace{-10pt}

\paragraph{Bone definition.} 
When modeling the object-level motion of general articulated objects, such as a stapler in \Cref{fig2}, a straightforward design is to follow the previous methods that model the motion of humans and animals from videos. In this design, the motion of a stapler is considered analogous to the arm of a human, and the rig is defined on three joints to control the stapler's motion. The joints will be optimized to align with the positions at the ends of the object's parts to minimize energy as illustrated in the middle of \Cref{fig2} (PPR).  

Applying PPR \cite{yang2023ppr} in such a design results in noticeable artifacts like bending shapes and corrupted motion, which is unacceptable for objects characterized by multiple rigid components. To address this limitation, we propose to define the rig on the bones, ideally the part centroids as illustrated in \Cref{fig2} (Ours). Consequently, each rigid part is strongly associated with one bone, effectively defining the motion of the articulated objects. The number of bones, denoted as $B$, depends on the number of rigid components in an articulated object.

\vspace{-13pt}

\paragraph{Skinning weights.}
We define the skinning weights as $\mathbf{W} = \{w_{0},..., w_{B-1}\} \in \mathbb{R}^{B}$. Given a 3D point $\mathbf{X}$, we compute the Gaussian skinning weights \cite{yang2022banmo, song2023moda} based on the Mahalanobis distance $d_M(\mathbf{X})$ between 3D points and the Gaussian bones, 
\begin{equation}
    d_{M}(\mathbf{X}) = (\mathbf{X}-\mathbf{O})^{T}\mathbf{V}^{T}\boldsymbol{\Lambda}^{0}\mathbf{V}(\mathbf{X}-\mathbf{O}),
\end{equation}
where $\mathbf{O} \in \mathbb{R}^{B \times 3}$ are bone centers, $\mathbf{V} \in \mathbb{R}^{B \times 3 \times 3}$ are bone orientations and $\boldsymbol{\Lambda}^{0} \in \mathbb{R}^{B \times 3 \times 3}$ are diagonal scale matrices. Each Gaussian bone has three parameters for
center, orientation, and scale respectively, which are all optimized during training. To further refine the Gaussian skinning weights, we incorporate delta skinning weights learned by an MLP,

\begin{equation}
    \mathbf{W} = \mathrm{softmax}(d_M(\mathbf{X}) + \mathbf{W}_{\Delta}),
    \label{weight}
\end{equation}
where $\mathbf{W}_{\Delta} = \mathbf{MLP}_{\mathrm{skin}}(\mathbf{X}_\mathrm{bone})$ is the delta skinning weights. $\mathbf{X}_\mathrm{bone} \in \mathbb{R}^{B \times 3}$ denotes the relative positions of point $\mathbf{X}$ in the bone coordinates.

However, given that general articulated objects are typically piece-wise rigid, the refined Gaussian skinning weights $\mathbf{W}$ may introduce redundant associations to multiple bones for each point, thus hampering the rigidity of the parts. Therefore,  we aim to make the skinning weights quasi-sparse to minimize the influence of other bones and ensure a strong association with their corresponding bone for 3D points. We first introduce a temperature factor $\gamma$ to the calculation of $\mathbf{W}$ to stimulate the sparsity,

\begin{equation}
    \mathbf{W}^{s} = \mathrm{softmax}(\frac{d_M(\mathbf{X}) + \mathbf{W}_{\Delta}}{\gamma}).
    \label{weight}
\end{equation}
\vspace{-20pt}

\paragraph{Geodesic point assignment.} We further propose a geodesic point assignment process to help correctly assign each point to the corresponding bone or joint, hence preventing surface tearing and corrupted motion. We can further enhance the sparsity of the skinning weights for the points in the rigid parts while keeping the weights unchanged for points in the joints based on the assignment, therefore, achieving quasi-rigid blend skinning.

The accuracy of Mahalanobis distance calculation depends on the precision of bone properties, including center, orientation, and scale. However, since these properties are all optimized during training, uncertainties are introduced into the calculation process. This can lead to inaccurate point assignments, as observed in our experiments. For instance, a point near the surface of one component may have a shorter Mahalanobis distance to a bone belonging to a different component.

\begin{figure}
  \centering
  \setlength{\abovecaptionskip}{0.cm}
  \includegraphics[scale=0.35]{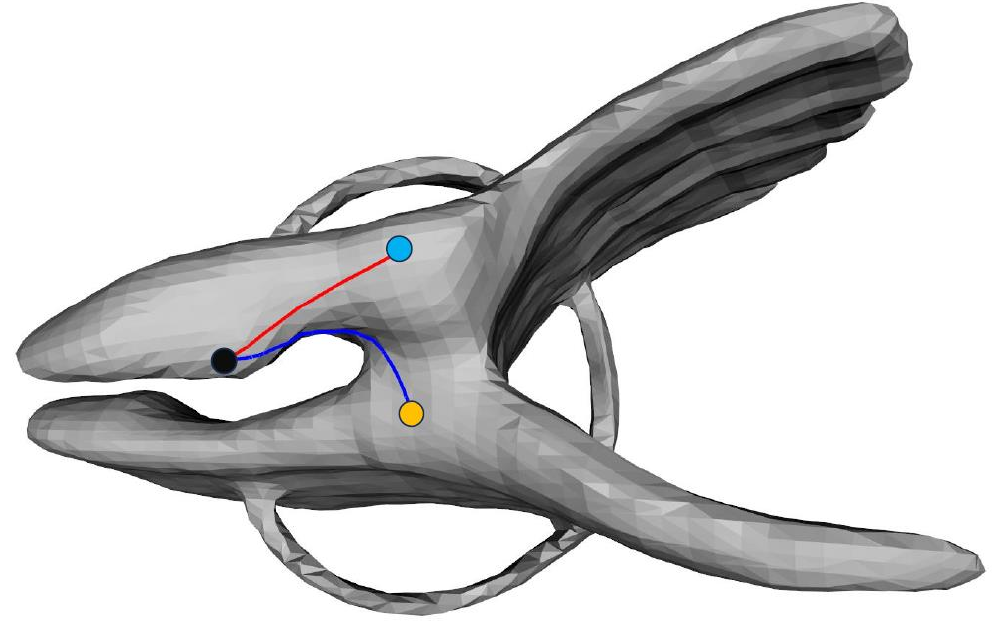}
  \caption{
  \textbf{Geodesic distances between 3D point and bones.} Geodesic distance can correctly associate the 3D point (black) with the top bone (blue) rather than the bottom bone (yellow) by following the shortest path on the mesh surface. Shorter distances indicate stronger associations.
  } 
  \label{fig4}
  \vspace{-15pt}

\end{figure}

Additionally, in cases where a point exhibits similar Mahalanobis distances to multiple bones, determining whether the point is associated with joints or rigid components remains challenging. To address these issues, we employ geodesic distance as depicted in \Cref{fig4}, which provides a measure of the shortest path between two points along a mesh surface.

To elaborate, for a point $\mathbf{X}$, we initially set up a point assignment vector $\mathbf{M}=\mathbf{0} \in \mathbb{R}^{B}$, representing the assignment of the point to its corresponding bone. We proceed by identifying the nearest bone $b_i$ and the second nearest bone $b_j$ to $\mathbf{X}$, based on their Mahalanobis distances $d_M^i$ and $d_M^j$. Given that geodesic distance calculations require a mesh surface, we first extract a canonical mesh using the marching cubes algorithm \cite{marchingcube}. Following this, we employ the KNN algorithm to locate the nearest vertices $\mathbf{\hat{X}}$, $\hat{b}_i$, and $\hat{b}_j$ relative to the point $\mathbf{X}$ and the centers of bones $b_i$ and $b_j$. Finally, we compute the geodesic distances $d_G^i$ and $d_G^j$ from $\mathbf{\hat{X}}$ to $\hat{b}_i$ and $\hat{b}_j$, respectively, utilizing the exact geodesic algorithm as described in \cite{mitchell1987discrete}. 

\renewcommand{\algorithmicrequire}{\textbf{Input:}}
\renewcommand{\algorithmicensure}{\textbf{Output:}}
\begin{algorithm}[t]
\caption{Geodesic point assignment}
\label{algorithm1}
\begin{algorithmic}[1]
\Require   Point assignment $\mathbf{M}=\mathbf{0} \in \mathbb{R}^{B}$, Mahalanobis distance $d_M^i$ and $d_M^j$, geodesic distance $d_G^i$ and $d_G^j$, bone index $i$ and $j$, hyperparameters $\eta$, $\zeta$.
\Ensure  Updated assignment $\mathbf{M}$

\If{$d_M^i/d_M^j < 1-\eta$}
    \State $\mathbf
    {M}[i] \gets 1$
\ElsIf{$\frac{|d_G^i-d_G^j|}{\min(d_G^i, d_G^j)} < \zeta$}
    \State $\mathbf{M}[i], \mathbf{M}[j] \gets \mathbf{1}$ \Comment{Assigning to joints}
\Else
    \State $\mathbf{M}[\text{argmin}(d_G)] \gets 1$ 
\EndIf

\end{algorithmic}

\end{algorithm}

As illustrated in \Cref{algorithm1}, if the ratio of Mahalanobis distances $d_M^i/d_M^j$ is less than $1-\eta$, which means the point is obviously closer to $b_i$, we assign a value of 1 to the $i^{th}$ element of $\mathbf{M}$. If the point is close to both bones, we check if $\frac{|d_G^i-d_G^j|}{\min(d_G^i, d_G^j)} < \zeta$, which means the geodesic distances are close. If the Mahalanobis distance and geodesic distance from the points to the bone $i$ and $j$ are both similar, we assign the value 1 to the $i^{th}$ and the $j^{th}$ elements of $\mathbf{M}$, as assigning the point to joints. If neither of the previous conditions is satisfied, we assign the point to the bone with the shortest geodesic distance. The distances are all passed through a softmax layer before being input into \Cref{algorithm1}.

As the mesh evolves during training, we refrain from applying the point assignment directly to the skinning weights as a mask. Instead, we impose penalties on the weights associated with bones not corresponding to the targeted point with a sparse skinning loss,

\begin{equation}
    \mathcal{L}_{sparse} = \frac{\sum \left\|\mathbf{W}^{s}\odot\mathbf{\bar{M}}\right\|^{2}  }{\sum \mathbf{\bar{M}}}, 
    \label{loss_w}
\end{equation}
where $\odot$ denotes Hadamard product and $\mathbf{\bar{M}}=1-\mathbf{M}$, since $\mathbf{M}$ indicates the correct assignment and we want to penalize the weights everywhere else. For points that have been assigned to joints, the skinning weights are not penalized. 

\vspace{-10pt}
\paragraph{Quasi-rigid blend skinning.}
With the learned quasi-sparse skinning weights, the articulated motion of 3D points under pose $\boldsymbol{\psi}_{p}$ can be obtained using our quasi-rigid blend skinning (QRBS):

\begin{equation}
    \mathbf{X(\boldsymbol{\psi}_{p})} = \mathbf{T}_{\mathrm{obj}}\mathbf{X} = (\sum_{b=0}^{B-1}{w_{b}^{S}\mathbf{T}_{b}})\mathbf{X},
    \label{quasi}
\end{equation}

where $\{\mathbf{T}_{0},...,\mathbf{T}_{B-1}\} = \mathbf{MLP}_{\mathrm{pose}}(\boldsymbol{\psi}_{p})$ and are represented by dual quaternions.
This articulated motion is invertible by inverting $\mathbf{T}_{b}$ in \Cref{quasi} and recomputing the skinning weights in \Cref{weight}. We utilize a 3D cycle loss \cite{yang2022banmo, li2021neural} to supervise this invertible process.
\vspace{-3pt}
\subsection{Volume rendering and optimization}
\label{loss}
\vspace{-3pt}
\paragraph{Volume rendering.}
We use the volume rendering in NeRF \cite{mildenhall2021nerf} to synthesize images. With the pixel location $\mathbf{x}^{t} \in \mathbb{R}^{2}$, the $n$-th sampled point along the ray that originates from $\mathbf{x}^{t}$ is $\mathbf{X}_{n}^{t}$. 
The color and opacity are given by:
\begin{equation}
    \mathbf{c}(\mathbf{x}^{t}) =\sum_{n=1}^{N}\tau_{n}\mathbf{c}_{n}^{t},
\quad
    \mathbf{o}(\mathbf{x}^{t}) = \sum_{n=1}^{N}\tau_{n},
\end{equation}
where $\tau_{n}=\alpha_{n}{\prod_{m=1}^{n-1}(1-\alpha_{m})}$, $\alpha_{n} = 1 - \mathrm{exp}(-\sigma_{n}\delta_{n})$, $N$ is the number of sampled points, $\delta_{n}$ is the distance between the $n$-th point and the next, and $\sigma_{n}$ is the density in \Cref{density}.

\vspace{-10pt}
\paragraph{Optimization.}
Except for the sparse skinning loss in \Cref{loss_w}, we optimize our models with multiple reconstruction losses (color, object mask, optical flow, pixel features) that are similar to existing methods \cite{yang2022banmo, song2023moda, yang2023ppr}. These losses are employed to minimize the difference between the predicted results and the observed ones, alongside regularization terms.
\begin{equation}
\vspace{-5pt}
    \mathcal{L} = \mathcal{L}_{rgb} + \mathcal{L}_{mask} + \mathcal{L}_{flow} + \mathcal{L}_{feature} + \mathcal{L}_{sparse} + \mathcal{L}_{reg}.
\end{equation}

The predicted mask, optical flow, and pixel features are obtained from off-the-shelf methods \cite{yang2023track, yang2019volumetric, oquab2023dinov2}. Please refer to the supplementary materials for the regularization terms.
\vspace{-10pt}
\section{Experiments}
\vspace{-5pt}
\subsection{Dataset, metrics, and implementation details}
\vspace{-5pt}
\paragraph{Real-world videos.}

To demonstrate the effectiveness of REACTO, we conducted evaluations on real-world videos with only partial views of different articulated objects such as laptops, staplers, scissors, faucets, nail clippers, glasses, and more. These videos were captured using a phone camera with no control over camera movements. For detailed information about the videos, please refer to the supplementary materials. 
In the preprocessing stage, we employed methodologies outlined in Lab4D \cite{lab4d}. Specifically, we utilized Track Anything \cite{yang2023track} for predicting object silhouettes, VCN-robust \cite{yang2019volumetric} for optical flow estimation, and DINOv2 \cite{oquab2023dinov2} for extracting pixel features. Additionally, we also annotate sparse camera poses (approximately 4 annotations per video) for camera estimation. These annotations serve as initialization and will be further optimized during the training process. To distinguish from synthetic data, we prepend \textit{real-} to articulated objects (e.g., \textit{real-laptop}).

\vspace{-10pt}
\paragraph{Synthetic videos.}
To evaluate our method quantitatively, we render videos using PartNet-Mobility dataset \cite{Xiang_2020_SAPIEN, chang2015shapenet, Mo_2019_CVPR} that provide ground truth meshes. We chose 3 categories for evaluation in this paper, namely \textit{USB}, \textit{stapler}, and \textit{scissors}. For more results from other categories, please refer to the supplementary material.
For each articulated object, we render 100 frames with the camera moving through a 120-degree azimuthal angle and a 30-degree polar angle using Blender \cite{blender2022}. The sequence consists of 50 frames corresponding to 50 consecutive articulations, followed by another 50 frames in reverse order. We train the synthetic dataset with ground truth object silhouettes. Similar to the process for real-world videos, we utilize VCN-robust \cite{yang2019volumetric} and DINOv2 \cite{oquab2023dinov2} for predicting optical flow and pixel features, respectively. The initial camera poses are obtained in the same manner as those for real-world videos. \\

% \vspace{-15pt}
\paragraph{Metrics.}
To quantitatively evaluate various methods, we employ Chamfer distance (CD) \cite{fan2017point} and F-scores as our metrics. For CD, lower values indicate better performance. F-scores are compared across different methods at distance thresholds $d = 10\%$ and $d = 5\%$. A higher F-score is better. As the ground truth meshes from PartNet-Mobility dataset \cite{Xiang_2020_SAPIEN} exhibit limited vertices and uneven distribution, we uniformly sample $10,000$ points using PyTorch3D \cite{ravi2020pytorch3d} from both predicted and ground truth meshes to compute Chamfer Distance (CD) and F-scores, which ensures a fair and robust evaluation.

% \vspace{-10pt}
\paragraph{Implementation details.}
We employ the AdamW optimizer to optimize the model for 4,000 iterations. For all objects, we start with the same shape of a unit sphere as PPR \cite{yang2023ppr}. The reconstructed meshes are extracted using marching cubes on a $128^{3}$ grid. For additional implementation details, please refer to the supplementary materials. 

\subsection{Comparison with state-of-art methods}
\label{seccompare}
\paragraph{Baselines.} We compare our method with BANMo \cite{yang2022banmo}, MoDA \cite{song2023moda} and PPR \cite{yang2023ppr}. These methods were originally designed for modeling humans or animals from videos. For the deformation model, BANMo employs linear blend skinning, while MoDA and PPR utilize dual quaternion blend skinning (note that the learning of dual quaternion differs between MoDA and PPR). 

To ensure fair comparisons, we report the results of BANMo, MoDA, and PPR with rigging on bones in this section. Each method utilizes 64 sampled points per ray to ensure consistent evaluation conditions. We supply BANMo, MoDA, and PPR with the same initial camera poses.

\begin{figure*}
  \centering
  \setlength{\abovecaptionskip}{0.cm}
  \includegraphics[scale=0.435]{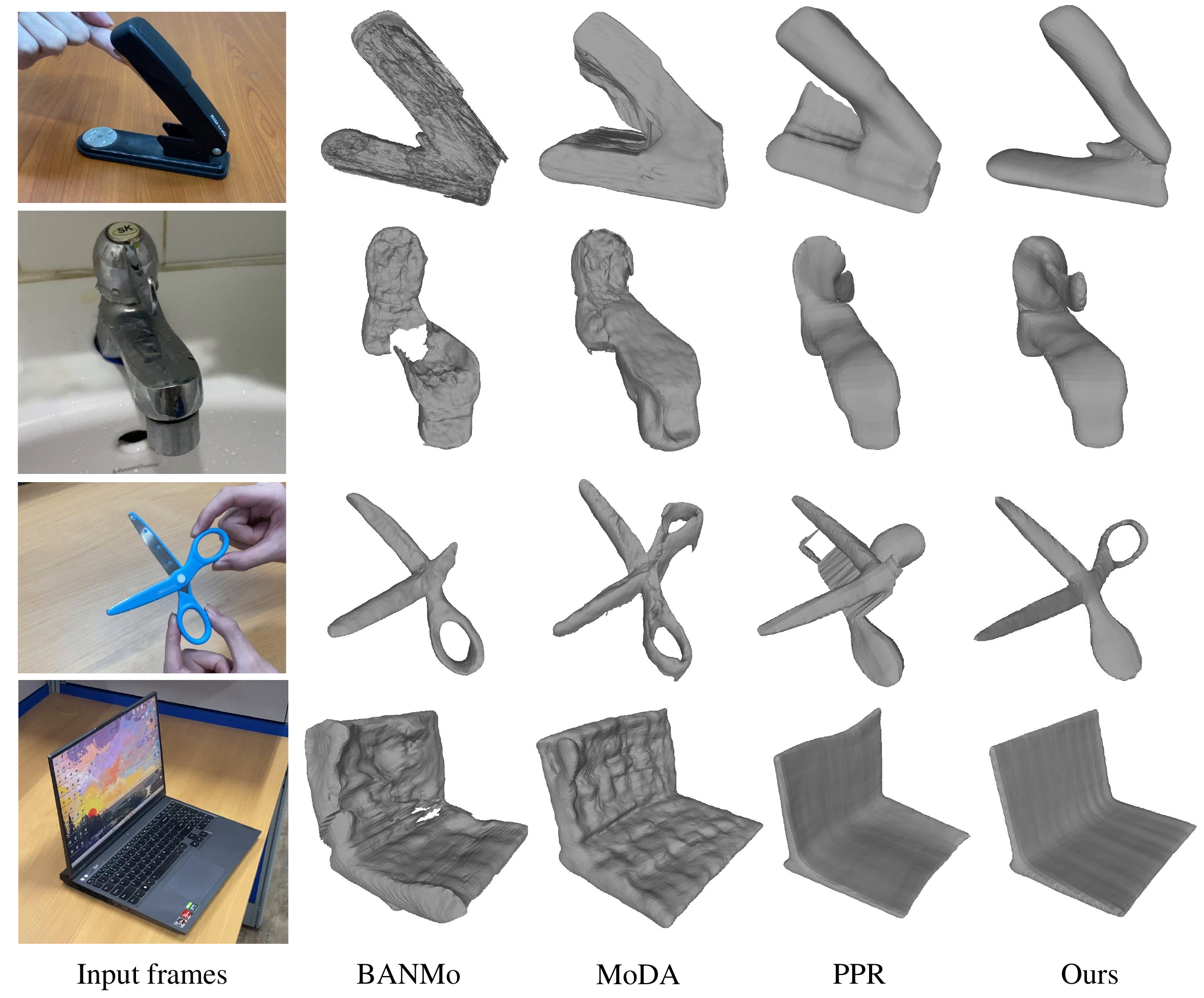}
  % \vspace{5pt}
  \caption{\textbf{Qualitative comparison of our method with BANMo \cite{yang2022banmo}, MoDA \cite{song2023moda} and PPR \cite{yang2023ppr}.} BANMo and MoDA struggle with complete shape reconstruction (\textit{real-faucet}, \textit{real-scissors}). Non-smooth surfaces (BANMo on \textit{real-stapler}, MoDA on \textit{real-scissors}, BANMo and MoDA on \textit{real-laptop}) are also observed. The results of PPR are smoother but with surface tearing (\textit{real-stapler}, \textit{real-scissors}), over-smoothed joints (\textit{real-faucet}, \textit{real-laptop}, \textit{real-stapler}), and inaccuracies in motion modeling (\textit{real-faucet}, \textit{real-laptop}). In contrast, REACTO outperforms these methods, excelling in the shape and deformation reconstruction of articulated objects. Please find the video results in the supplementary material.
  } 
  \label{fig_compare}
  % \vspace{-10pt}
\end{figure*}

% \vspace{-10pt}
\paragraph{Results.} The qualitative and quantitative results are presented in \Cref{fig_compare} and \Cref{tab_compare}, respectively. In \Cref{fig_compare}, both BANMo and MoDA struggle to reconstruct the complete shape of articulated objects. This is evident in instances such as both methods facing difficulties with \textit{real-faucet} and BANMo also encountering challenges with \textit{real-scissors}. These two methods often yield non-smooth surfaces, as observed with BANMo on \textit{real-stapler}, MoDA on \textit{real-scissors}, and both methods on \textit{real-laptop}. Although PPR generates smoother surfaces compared to BANMo and MoDA, it still encounters challenges in accurately modeling the motion of articulated objects. Notably, it introduces surface tearing artifacts in cases such as \textit{real-stapler} and \textit{real-scissors}. Also, we observe over-smoothed joints in \textit{real-faucet}, \textit{real-stapler}, and \textit{real-laptop}. Furthermore, when applied to \textit{real-faucet} and \textit{real-laptop}, PPR demonstrates inaccuracies in modeling motions, such as the rotation of the handle in \textit{real-faucet} and the folding motion of \textit{real-laptop}. In contrast, our REACTO consistently outperforms these methods, with superior capabilities in modeling the shape and deformation of various articulated objects.

Our quantitative results support qualitative observations, demonstrating that REACTO outperforms all baselines across all metrics on the synthetic data.

\begin{table*}[t]
  \centering
  \caption{\textbf{Quantitative comparison between different methods.} Our method has better performance than BANMo \cite{yang2022banmo}, MoDA \cite{song2023moda}, and PPR \cite{yang2023ppr} across all metrics.}
  \label{tab_compare}
  \begin{tabular}{cccccccccc}
    \toprule
    \multirow{2}{*}{Method} & \multicolumn{3}{c}{USB} &  \multicolumn{3}{c}{stapler} & \multicolumn{3}{c}{scissors} \\ \cmidrule{2-10}
  &  CD($\downarrow$) & F(10\%, $\uparrow$) & F(5\%, $\uparrow$) &CD($\downarrow$) &  F(10\%, $\uparrow$) &  F(5\%, $\uparrow$) &CD($\downarrow$)& F(10\%, $\uparrow$) & F(5\%, $\uparrow$)  \\
    \midrule
     BANMo  & 20.3  & 65.1 & 45.0 & 19.1 & 57.8 &32.8  & 19.9 & 66.8 & 41.4 \\
    MoDA &  17.1 & 74.9 & 49.5 & 18.8 & 64.2 & 40.3 & 14.8 & 77.7 & 42.3\\
      PPR & 20.7  & 65.7 & 38.9 & 16.8 & 67.5 & 40.0  & 16.1 & 71.4 & 39.9 \\
      Ours   &  \textbf{15.3} & \textbf{78.6} & \textbf{51.5}& \textbf{14.3} & \textbf{75.5} & \textbf{42.7}   & \textbf{14.0} & \textbf{78.2} & \textbf{43.9}    \\
    \bottomrule
  \end{tabular}
\end{table*}

\begin{figure}
  \centering
  \setlength{\abovecaptionskip}{0.cm}
  \includegraphics[scale=0.24]{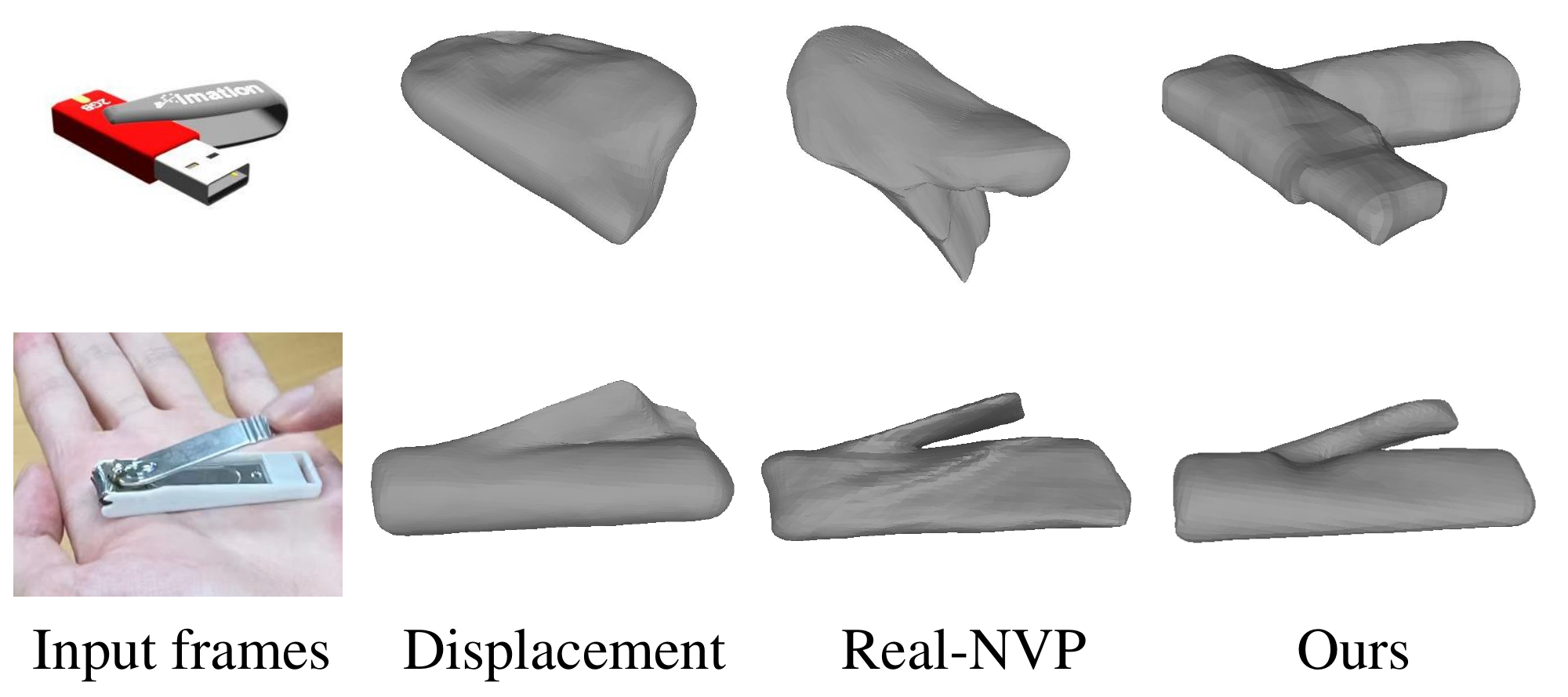}
  \caption{\textbf{Ablation study on deformation models.} 
  We compare displacement field \cite{wei2022self} and Real-NVP \cite{dinh2016density, lei2022cadex} with our QRBS on synthetic \textit{USB} and \textit{real-nail clipper}. The displacement field struggles to accurately separate the motion of the two rigid parts in both \textit{USB} and \textit{real-nail clipper}. Real-NVP also fails to separate the two rigid parts of \textit{USB} and produces non-smoothness when modeling the motion of \textit{real-nail clipper}. In contrast, our QRBS consistently outperforms both methods in both cases.
  } 
  \label{fig_disnvp}
  \vspace{-15pt}
\end{figure}
\vspace{-3pt}
\subsection{Ablation study on deformation models}
\vspace{-2pt}
\begin{figure}
  \centering
  \setlength{\abovecaptionskip}{0.cm}
  \includegraphics[scale=0.27]{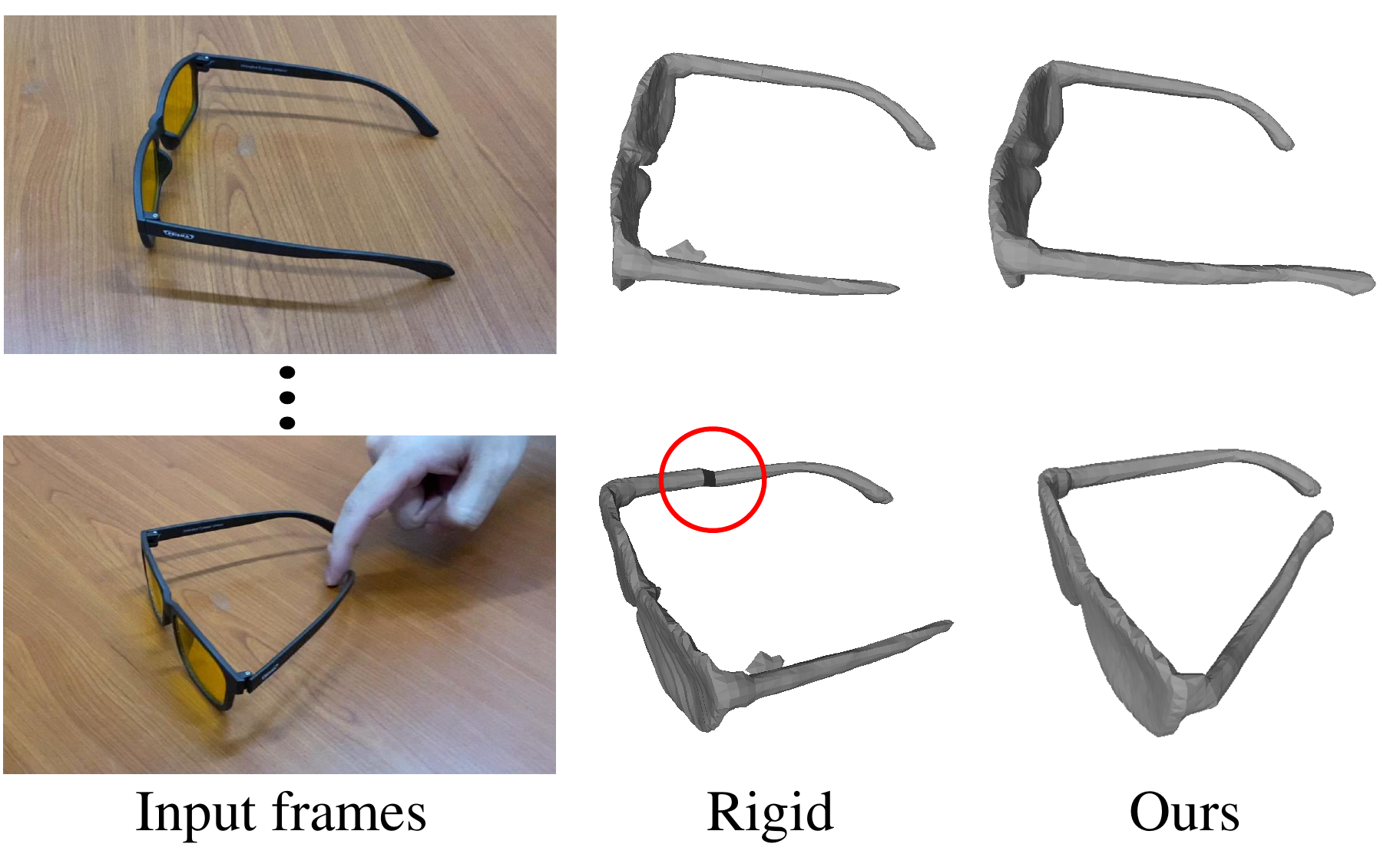}
  \caption{\textbf{Rigid skinning vs. Quasi-rigid blend skinning.} For rigid skinning, we binaryize skinning weights by setting the largest $w_b$ to 1 and all others to 0, which fails to model the articulation while causing seam artifacts on the leg of \textit{real-glasses} (in the red circle).
  } 
  \label{fig_hard}
  \vspace{-15pt}
\end{figure}

\begin{table*}
  \centering
  \caption{\textbf{Quantitative ablation studies on deformation models.} Our method outperforms the displacement field, Real-NVP, and rigid skinning across various data.}
  \label{table_ablation}
  \begin{tabular}{cccccccccc}
    \toprule
    \multirow{2}{*}{Method} & \multicolumn{3}{c}{USB} &  \multicolumn{3}{c}{stapler} & \multicolumn{3}{c}{scissors} \\ \cmidrule{2-10}
  &  CD($\downarrow$) & F(10\%, $\uparrow$) & F(5\%, $\uparrow$) &CD($\downarrow$) &  F(10\%, $\uparrow$) &  F(5\%, $\uparrow$) &CD($\downarrow$)& F(10\%, $\uparrow$) & F(5\%, $\uparrow$)  \\
    \midrule
     Displacement  & 19.7   & 59.5 & 28.2 & 17.9 & 63.8 & 30.8 & 19.1 & 58.6 & 30.9 \\
    Real-NVP &  17.6  & 70.7 & 47.2 & 16.0 & 70.4 & 32.6 & 19.6 & 63.6 & 32.7 \\
      Rigid & 16.3 & 73.5 & 49.3 & 15.1 & 72.8& 41.1  & 14.8 & 76.4 & 43.7 \\
      Ours   &  \textbf{15.3} & \textbf{78.6} & \textbf{51.5}& \textbf{14.3} & \textbf{75.5} & \textbf{42.7}  & \textbf{14.0} & \textbf{78.2} & \textbf{43.9}     \\
    \bottomrule
  \end{tabular}
\end{table*}

In this section, we compare our quasi-rigid blend skinning with other deformation models employed for articulated object motion, such as displacement field in NASAM \cite{wei2022self} and invertible Real-NVP \cite{dinh2016density} in CaDeX \cite{lei2022cadex}. 

The qualitative and quantitative results are presented in \Cref{fig_disnvp} and \Cref{table_ablation}, respectively. For the synthetic \textit{USB}, both displacement field and Real-NVP struggle to accurately distinguish the motion of the two rigid parts. In contrast, our method successfully models the motion with the optimized rigging system. For \textit{real-nail clipper}, the displacement field still fails to separate the two rigid parts. Real-NVP introduces non-smoothness during motion, while our method maintains a consistently smooth mesh surface. The quantitative results on synthetic data further confirm that our quasi-rigid blend skinning offers a more reasonable approach than other deformation models for modeling the motion of general articulated objects.

Besides, we also propose a straightforward design for rigid skinning. For the skinning weights $w_{b}, b \in [0, B-1]$ ($w_b \in [0, 1], \sum_{b}w_{b} = 1$), we binaryize them by setting the largest $w_b$ to 1 and all others to 0. As illustrated in \Cref{table_ablation}, rigid skinning exhibits comparable performance with our method when evaluated on synthetic data. However, it may lead to seam artifacts, as exemplified in \Cref{fig_hard}, particularly noticeable in the leg of the glasses.

\section{Conclusion}

In this paper, we introduce REACTO, a groundbreaking method for reconstructing general articulated 3D objects from single casual videos, achieving enhanced modeling and precision by redefining rigging structures and employing Quasi-Rigid Blend Skinning. QRBS ensures the rigidity of each component while retaining smooth deformation on the joints by utilizing quasi-sparse skinning weights and geodesic point assignment. Extensive experiments show that REACTO outperforms existing methods in fidelity and detail on both real and synthetic datasets. 

\textbf{Limitations:} As casual videos typically offer only partial views of objects, the quality of surface reconstruction may suffer on the unseen side. The limitations of this approach will be further detailed in the supplementary materials.

\section*{Acknowledgements} 

This research work is supported by the Agency for Science, Technology and Research (A*STAR) under its MTC Programmatic Funds (Grant No. M23L7b0021).

{
    \small
    \bibliographystyle{ieeenat_fullname}
    \bibliography{main}
}


\end{document}